\documentclass[10pt,twocolumn,letterpaper]{article}

\usepackage[pagenumbers]{iccv} % To force page numbers, e.g. for an arXiv version

\usepackage{multirow}
\usepackage{booktabs}
\usepackage{graphicx} % For checkmarks
\usepackage{array}
\usepackage{multirow}
\usepackage{makecell} % Fix multi-line text inside multirow
\usepackage{comment} % Enable block comments
\usepackage{algorithm}
\usepackage{algpseudocode}
\usepackage{tcolorbox}
\usepackage{capt-of}
\usepackage{colortbl}  % Add in the preamble
\usepackage{xcolor}     % For defining colors
\usepackage{pifont}
\newcommand{\cmark}{\textcolor{green}{\ding{51}}}
\newcommand{\xmark}{\textcolor{red}{\ding{55}}}
\usepackage{multicol}
\definecolor{iccvblue}{rgb}{0.21,0.49,0.74}
\usepackage[pagebackref,breaklinks,colorlinks,allcolors=iccvblue]{hyperref}

\def\paperID{3935} % *** Enter the Paper ID here
\def\confName{ICCV}
\def\confYear{2025}

\title{Unlocking Pretrained LLMs for Motion-Related Multimodal Generation: A Fine-Tuning Approach to Unify Diffusion and Next-Token Prediction}

\author{
  Shinichi Tanaka\textsuperscript{$\dagger$}\thanks{Equal contribution.} \quad\quad 
  Zhao Wang\textsuperscript{$\dagger$, $\ddagger$}\footnotemark[1] \quad\quad 
  Yoichi Kato\textsuperscript{$\dagger$} \quad\quad 
  Jun Ohya\textsuperscript{$\dagger$}\\[1ex]
  \textsuperscript{$\dagger$} Waseda University \quad\quad 
  \textsuperscript{$\ddagger$} Sony Group Corporation
}

\begin{document}
\maketitle
% Resume normal two-column layout
\begin{abstract}

%For motion-related multimodal generation tasks, diffusion models generate high-quality human motions but lack generalizability, limiting their effectiveness in downstream tasks such as motion captioning. In contrast, LLM-based models offer flexible modality control but struggle with cross-modal distribution learning, leading to lower generation quality. 

In this paper, we propose a unified framework that leverages a single pretrained LLM for \textbf{Mo}tion-related \textbf{Mu}ltimodal \textbf{G}eneration, referred to as \textit{MoMug}. \textit{MoMug} integrates diffusion-based continuous motion generation with the model's inherent autoregressive discrete text prediction capabilities by fine-tuning a pretrained LLM. This enables seamless switching between continuous motion output and discrete text token prediction within a single model architecture, effectively combining the strengths of both diffusion- and LLM-based approaches. Experimental results show that, compared to the most recent LLM-based baseline, \textit{MoMug} improves FID by 38\% and mean accuracy across seven metrics by 16.61\% on the text-to-motion task. Additionally, it improves mean accuracy across eight metrics by 8.44\% on the text-to-motion task. To the best of our knowledge, this is the first approach to integrate diffusion- and LLM-based generation within a single model for motion-related multimodal tasks while maintaining low training costs. This establishes a foundation for future advancements in motion-related generation, paving the way for high-quality yet cost-efficient motion synthesis.%\footnote{Code will be publicly available in the camera-ready version.}

\end{abstract}    
\section{Introduction}
\label{sec:intro}

\begin{figure}[ht]
    \centering
    \includegraphics[width=0.48\textwidth, trim=160 397 160 0, clip]{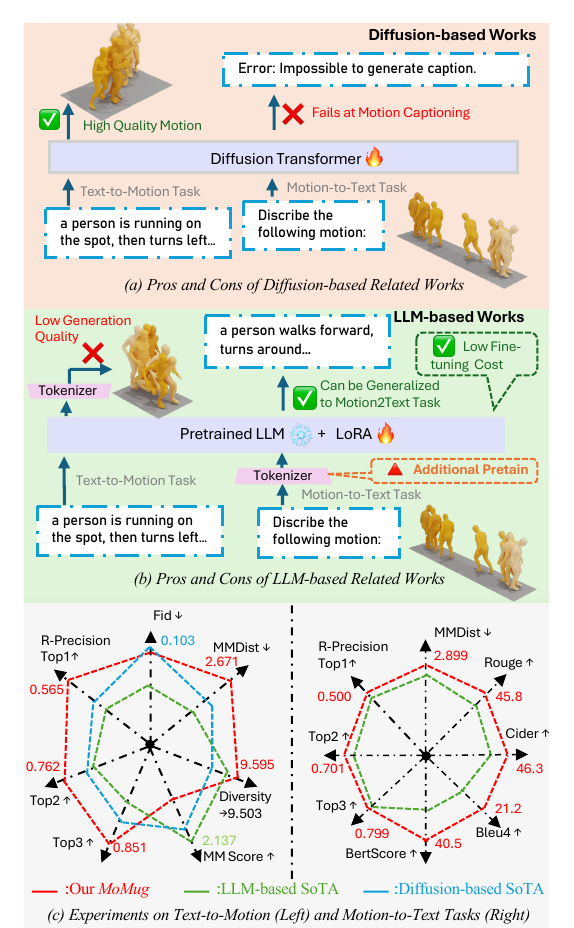} % Replace with your file name and extension
    \vspace{-20pt} % adjust the negative space here
    \caption{(a) and (b) illustrate the pros and cons of diffusion-based and LLM-based approaches, which motivate us to leverage a pre-trained LLM to take advantage of both. As shown in (c), our \textit{MoMug}—the first work to unify both approaches in a single pre-trained LLM—significantly improves performance in both text-to-motion and motion-to-text tasks across multiple metrics.}
    \label{fig:teaser}
\end{figure}

Motion-related multimodal generation tasks, including text-to-motion (motion generation) \citep{guo2022generating}, motion-to-text (motion captioning) \citep{guo2022tm2tstochastictokenizedmodeling, radouane2023guided}, and motion editing \citep{athanasiou2024motionfix}, are crucial for AI and human-computer interaction. These tasks enable machines to generate, understand, and describe human motions, benefiting applications such as film, video games, AR/VR, robotics, and digital humans by enhancing realism and interactivity \citep{zhu2023human}.
\definecolor{verylightgray}{gray}{0.95}
\definecolor{headergray}{gray}{0.85}  % Medium gray for header
\definecolor{rowlight}{gray}{0.96}    % Light gray for Diffusion-based
\definecolor{rowdark}{gray}{0.80}     % Darker gray for LLM-based
\definecolor{highlight}{RGB}{230, 242, 255} % Light blue for MoMug row

Recent research in motion-related tasks has followed two main trends. The first is diffusion-based methods \citep{ho2020denoising, 2205.11487}. Since diffusion models are trained to denoise from a Gaussian noise distribution to the target motion distribution \citep{dhariwal2021diffusion}, they inherently focus on maintaining sequence-wide consistency \citep{song2020denoising}. Compared to LLM-based methods \citep{jiang2024motiongpt, motiongptother2023, wang2024motiongpt2generalpurposemotionlanguagemodel}, which rely on discrete tokenization, diffusion models use DDPM loss, operating in a continuous space. This key distinction allows diffusion models to generate smoother and more natural motion sequences, as they inherently model continuous motion dynamics rather than discrete transitions between tokens \citep{zhang2023mld, liu2023remodiffuse}. Similar advantages have been observed in other generative fields, such as image \citep{ho2020denoising}, video \citep{ho2022video}, and audio generation \citep{huang2023make}. However, these models also have significant drawbacks: they are limited to generating only the motion modality \citep{chen2024motionllm, motiongptother2023}, as diffusion models are primarily designed for generating continuous data (e.g., images, audio, and motion). This limitation makes it impossible to generalize them to other motion-related tasks, such as motion-to-text.

The second trend leverages pretrained LLMs \citep{grattafiori2024llama3herdmodels, qwen2023} to generate motion and text within a single model. These approaches \citep{motiongptother2023, chen2024motionllm, wang2024motiongpt2generalpurposemotionlanguagemodel} benefit from the strong world knowledge \citep{zhang2023large} and generalization ability of LLMs, enabling them to produce both motion sequences and textual descriptions in a shared representation space. However, despite their flexibility, these methods have a major limitation: they tokenize motion into a discrete feature space and train the model using a next-token prediction approach (one-after-one supervised fine-tuning), as is common for text. Due to the autoregressive nature of this approach and its reliance on discrete representations \citep{zhang2022survey}, these models struggle to maintain the global coherence and continuity required for high-quality motion generation. As a result, they often underperform compared to diffusion-based models \citep{liu2023remodiffuse}, especially when evaluated using metrics that assess generation quality over distribution, such as Multi Modality Distance (MMDist) and Fréchet Inception Distance (FID), which measure the diversity and realism of generated motion.

Inspired by recent Multimodal LLM (M-LLM) methods in vision \citep{zhou2024transfusionpredicttokendiffuse} and video generation \citep{lian2023llmgroundedvideo}, as illustrated in Figure \ref{fig:teaser}, we are compelled to ask the following research question:

\textit{Can we leverage both diffusion and LLMs in a single unified model that can read and generate text while seamlessly switching to a diffusion-based approach for motion generation, all while maintaining low training costs?}

To answer this question, in this paper, we propose a unified framework that leverages a single pretrained LLM for \textbf{Mo}tion-related \textbf{Mu}ltimodal \textbf{G}eneration, referred to as \textit{MoMug}. Unlike existing methods that specialize in either diffusion-based continuous motion generation or discrete text and motion prediction, \textit{MoMug} seamlessly integrates both capabilities within a single LLM model. The model is trained with exposure to both modalities and their corresponding loss functions. By leveraging the shared representation space of an LLM, \textit{MoMug} ensures better alignment between motion and text, making it adaptable to various motion-related tasks, including text-to-motion generation, motion captioning, and motion editing. 

We summarize the key novelties of \textit{MoMug} as follows:

\begin{itemize}
    \item \textbf{Unified Motion-Text Generation:} \textit{MoMug} is the first framework to integrate diffusion-based continuous motion generation within a single LLM model for motion-related multimodal generation tasks, ensuring that it leverages the advantages of both diffusion and LLM-based methods mentioned above.
    
    \item \textbf{Efficient Fine-Tuning Approach:} Unlike general diffusion-integrated M-LLM methods that require training from scratch, \textit{MoMug} leverages pretrained LLMs and applies LoRA-based fine-tuning, significantly reducing computational costs while maintaining high performance.
    
    \item \textbf{Superior Performance and Generalization:} Experimental results demonstrate that, \textit{MoMug} improves FID by 38\% and mean accuracy across seven metrics by 16.61\% on the text-to-motion task. Additionally, it improves mean accuracy across eight metrics by 8.44\% on the text-to-motion task. Ablation studies confirm its adaptability across different LLM sizes and its strong \textit{generalization ability} across various motion-related tasks.
\end{itemize}

\section{Related Works}
\label{sec:related_work}

\paragraph{Human Motion Generation}
Human motion generation is a long-standing research problem that has been extensively studied using various approaches~\cite{zhang2022survey}. Recent advancements focus on motion generation with methods such as VAE-based models~\cite{petrovich2022temos}, graph-based techniques~\cite{yin2021graph}, BERT-based frameworks~\cite{guo2023momaskgenerativemaskedmodeling}, Flow Matching-based methods~\cite{hu2023motionflowmatchinghuman}, and state-space models~\cite{zhang2025motion}, making this field highly competitive and dynamic.

\paragraph{Diffusion based Methods.}  
Diffusion-based methods have become a leading approach for motion generation. These models gradually refine random noise into realistic motion sequences through an iterative denoising process~\citep{dhariwal2021diffusion}. This process ensures smooth and temporally consistent motion, which is crucial for text-to-motion tasks~\citep{song2020denoising}.

Several advancements have improved efficiency and flexibility. MDM~\citep{tevet2022human} applies diffusion directly to motion data, capturing rich temporal dependencies for text-conditioned generation. MLD~\citep{zhang2023mld} reduces computational cost by working in a learned latent space while maintaining quality. MoLA~\citep{mola2024} enhances motion generation and editing by combining latent diffusion with adversarial learning, supporting tasks like in-betweening and path-following. ReMoDiffuse~\citep{liu2023remodiffuse} improves motion refinement by integrating a retrieval mechanism that pulls similar examples from a motion database. These approaches highlight the effectiveness of diffusion models in creating natural, expressive, and controllable human motion.

\begin{figure*}[t]
   \centering
   \includegraphics[width=0.95\textwidth, trim=10 100 10 70, clip]{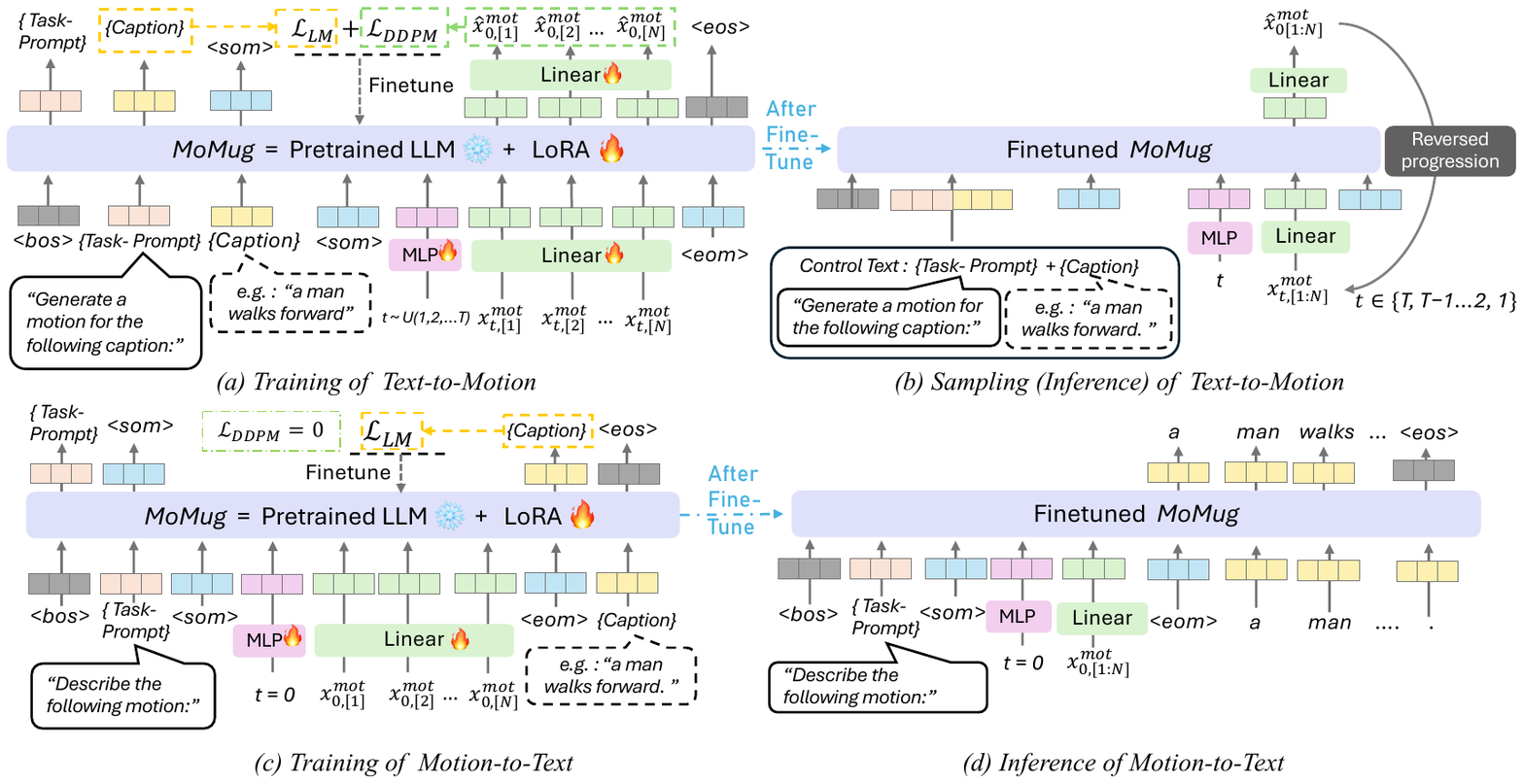} % Replace with your file name and extension
   \caption{Overview of \textit{MoMug}. The model consists of a pretrained LLM with LoRA fine-tuning and motion diffusion modeling, seamlessly switching between text-to-motion and motion-to-text modes. It processes mixed input sequences containing text tokens, the diffusion timestep \( t \), and motion frames \( \mathbf{x}^{\text{mot}}_{t,[1:N]} \). During training, in text-to-motion mode (a), the model optimizes both the language modeling loss \( \mathcal{L}_{\text{LM}} \) and the diffusion modeling loss \( \mathcal{L}_{\text{DDPM}} \), while in motion-to-text mode (c), \( \mathcal{L}_{\text{DDPM}} = 0 \). During inference, the model generates motion sequences via diffusion sampling (b) and text via LLM next-token prediction (d).} %(left) Text-to-Motion Mode: The model takes a task prompt (e.g., ``Generate a motion for the following caption:'') and a caption (e.g., ``a man walks forward'') as input, processes them through the LLM with LoRA adaptations, and generates motion tokens via MLP and Linear projections. (right) Motion-to-Text Mode: The model takes a task prompt (e.g., ``Describe the following motion:'') and motion sequence as input, processes them through the same architecture, and generates descriptive text captions for the given motion.}
   \label{fig:overview}
\end{figure*}

Despite their strengths, diffusion-based methods have some limitations. First, they are mainly designed for continuous motion generation and struggle with other tasks like text prediction in motion-to-text task. Second, they typically use CLIP~\citep{radford2021learning} as the text encoder, which limits their ability to fully utilize the broad language understanding of LLMs. Since CLIP focuses on vision-language alignment rather than deep text reasoning, these models may struggle with complex or unseen text descriptions, reducing their adaptability.
%-------------------------------------------------------------------------
\paragraph{LLM based Motion Muiltimodal Generation Methods.}

Recent advancements in LLM-based motion generation have demonstrated the effectiveness of treating human motion as a language. MotionGPT \cite{jiang2024motiongpt} introduces a framework that tokenizes 3D motion data, enabling various motion-related tasks such as generation, captioning, and prediction. Another version of MotionGPT \cite{motiongptother2023} fine-tunes LLMs as general-purpose motion generators using multimodal control signals. MotionGPT-2 \cite{wang2024motiongpt2generalpurposemotionlanguagemodel} further extends this approach with a versatile motion-language model that integrates textual and pose-based inputs for holistic motion synthesis. MotionLLM \cite{chen2024motionllm} explores motion understanding from both motion sequences and videos, reinforcing the role of LLMs in bridging language and motion. These works highlight the potential of LLMs in advancing human motion generation and comprehension.

However, a significant drawback of these LLM-based approaches is their reliance on tokenizing motion into a discrete feature space and employing a next-token prediction paradigm typical of text models. This autoregressive strategy, coupled with the limitations of discrete representations \citep{zhang2022survey}, undermines the ability to maintain global coherence and smooth continuity—both crucial for high-quality motion synthesis. Consequently, these methods often 
% lag behind 
outperformed by
diffusion-based models \citep{liu2023remodiffuse}, especially when evaluated using metrics like Multi Modality Distance (MMDist) and Fréchet Inception Distance (FID), which gauge the diversity and realism of the generated motion.

% typical of ~: ~に典型的な, ~に特有の
% undermine: 弱体化させる
% continuity: 連続bn
% gauge: 評価する

\section{Preliminaries}

\subsection{Diffusion Model for Human Motion Generation}
\label{sec:motion_modeling}

Diffusion models for motion synthesis involve a forward process that gradually adds Gaussian noise to raw motion data, and a reverse process that learns to denoise it. Here, we denote a continuous motion sequence $\mathbf{x}^{\text{mot}}_{t,[1:N]} \in \mathbb{R}^{d_\mathrm{motion}}$ with $N$ frames as $\mathbf{x}^{\text{mot}}_t$, where $t$ indicates the diffusion timestep and $d_\mathrm{motion}$ represents the dimensionality of the pose representation.

In the forward diffusion process, Gaussian noise is progressively added to the original motion data $\mathbf{x}^{\text{mot}}_0 \sim q(\mathbf{x}^{\text{mot}}_0)$ over $T$ timesteps, forming a Markov chain defined as: $q(\mathbf{x}_t|\mathbf{x}_{t-1}) := \mathcal{N}(\mathbf{x}_t; \sqrt{1-\beta_t}\mathbf{x}_{t-1}, \beta_t\mathbf{I})$, where $\beta_1, \dots, \beta_T$ represent a variance schedule. Following the notation from DDPM~\citep{ho2020denoising}, we define $\alpha_t := 1-\beta_t$ and $\bar{\alpha}_t := \prod_{s=1}^{t}\alpha_s$, simplifying the forward process to:
\begin{equation}
\label{eq:forward_diffusion}
\mathbf{x}_t = \sqrt{\bar{\alpha}_t}\mathbf{x}_0 + \sqrt{1-\bar{\alpha}_t}\boldsymbol{\epsilon},\quad \boldsymbol{\epsilon}\sim\mathcal{N}(\mathbf{0},\mathbf{I}).
\end{equation}

The reverse diffusion process aims to denoise $\mathbf{x}_t$ back into the clean motion data $\mathbf{x}_0$. Specifically for motion synthesis, MDM~\cite{tevet2022human} employs a neural network $G$, conditioned on timestep $t$ and a semantic textual prompt $txt$, to directly predict $\hat{\mathbf{x}}^{\text{mot}}_0$ from its noisy version $\mathbf{x}^{\text{mot}}_t$, where $\hat{\mathbf{x}}^{\text{mot}}_0$ is the prediction of the original motion data $\mathbf{x}^{\text{mot}}_0$. The corresponding objective function is given by:
\begin{equation}
\label{eq:ddpm}
\mathcal{L}_{\text{DDPM}} = \mathbb{E}_{\mathbf{x}^{\text{mot}}_{0},\boldsymbol{\epsilon},t}\left[\|\mathbf{x}^{\text{mot}}_{0} - G(\mathbf{x}^{\text{mot}}_{t}, t, txt)\|^2\right],
\end{equation}
where $t$ is uniformly sampled from $[1, T]$.

During sampling, motion sequences are generated by iteratively denoising from Gaussian noise $\mathbf{x}^{\text{mot}}_T$. At each step, $\mathbf{x}^{\text{mot}}_{t-1}$ is sampled from $\mathcal{N}(\mu_\theta(\mathbf{x}^{\text{mot}}_t, t, txt), \beta_t)$, where:
\begin{equation}
   \label{eq:reverse_diffusion}
   \small
   \begin{split}
       \mu_\theta(\mathbf{x}_t^{\text{mot}}, t, txt)
       &= \sqrt{\bar{\alpha}_t}\, G(\mathbf{x}_t^{\text{mot}}, t, txt) 
        + \sqrt{1 - \bar{\alpha}_t}\, \epsilon_\theta(\mathbf{x}_t^{\text{mot}}, t, txt),\\[1mm]
       \epsilon_\theta(\mathbf{x}_t^{\text{mot}}, t, txt)
       &= \left(\frac{\mathbf{x}_t^{\text{mot}}}{\sqrt{\bar{\alpha}_t}} - G(\mathbf{x}_t^{\text{mot}}, t, txt)\right)
       \sqrt{\frac{1}{\bar{\alpha}_t} - 1}\,.
   \end{split}
\end{equation}
In other words, this process predicts clean motion $\hat{\mathbf{x}}^{\text{mot}}_0 = G(\mathbf{x}^{\text{mot}}_t, t, txt)$ from noisy samples, then adds controlled noise to obtain $\mathbf{x}^{\text{mot}}_{t-1}$. The denoising continues until reaching $\mathbf{x}^{\text{mot}}_0$, the final motion sequence that aligns with the text prompt.

\subsection{Text Modality Modeling}
\label{sec:lang_modeling}

We adopt an autoregressive language modeling for text modality in \textit{MoMug}. We represent text as a sequence of discrete tokens: $\mathbf{x}^{\text{txt}} = (x^{\text{txt}}_1, x^{\text{txt}}_2, \ldots, x^{\text{txt}}_{N})$
where each $x^{\text{txt}}_i$ is a discrete token. Our language model parameterizes the joint probability over the text sequence as $P(\mathbf{x}^{\text{txt}}) = \prod_{i=1}^{N} P_\theta(x^{\text{txt}}_i \mid \mathbf{x}^{\text{txt}}{<i})$, where we parameterize the conditional probabilities using a neural network. The model is trained by minimizing the standard cross-entropy loss on textual tokens:
\begin{equation}
    \label{eq:lm}
    \mathcal{L}_{\text{LM}}
    = \mathbb{E}[-\log P_\theta(x^{\text{txt}}_i \mid \mathbf{x}^{\text{txt}}_{<i})].
\end{equation}

Importantly, the text modality component is shared across both the motion-to-text and text-to-motion tasks. This unified text modeling framework plays a crucial role in our proposed motion-related multimodality method, as it effectively bridges the gap between motion and language by leveraging a common representation.
% \subsection{Training Objective}
% \label{sec:training_obj}
% We follow Transfusion \cite{zhou2024transfusionpredicttokendiffuse} and jointly optimize the model for both motion diffusion and language modeling. Concretely, we combine the DDPM objective from MDM (Eq.\,2) with the language modeling loss (Eq.\,3):
% \begin{equation}
%     \label{eq:training_obj}
%     \mathcal{L}_{\text{total}}
%     = \lambda \, \mathcal{L}_{\text{LM}} + \mathcal{L}_{\text{DDPM}},
% \end{equation}
% where $\lambda$ is a weighting hyperparameter that balances the influence of text and motion objectives.
% TODO:
% tableの数値を埋める
% 8B の残りeopchを学習する
% edit task 実装
% 結果を整理する

\section{Methodology}
\label{sec:method}
%To handle a versatile motion-related miltimodal generation, we propose a unified motion-language framework named MoMug. 
As illustrated in Fig.~\ref{fig:overview}, \textit{MoMug} combines a pre-trained LLM with LoRA fine-tuning~\cite{hu2021lora} and motion diffusion modeling. This approach enables a single pretrained LLM to cost-effectively and seamlessly handle both the inverse diffusion process for human motion generation (Fig.~\ref{fig:overview}\textit{(b)}) and standard next-token prediction for motion-to-text (Fig.~\ref{fig:overview}\textit{(d)}) tasks, creating a versatile model for integrating diffusion-based motion generation with LLM-based text generation.

\subsection{Unified Sequence Respresentation}
We adopt Transformer~\cite{vaswani2023attentionneed} based LLM architecture~\cite{grattafiori2024llama3herdmodels} for motion-related multimodality generation including text-to-motion, motion-to-text tasks. 

We embed text $\mathbf{x}^{\text{txt}}$, diffusion time step $t$ and motion $\mathbf{x}^{\text{mot}}_0$ into LLM's hidden space $\mathbb{R}^d$ using the following formulation:

\begin{equation}
\label{eq:emb_text}
    \mathbf{h}^{\text{txt}} = \mathrm{Tokenizer}(\mathbf{x}^{\text{txt}})
\end{equation}
\begin{equation}
\label{eq:emb_t}
    \small
    \mathbf{h}^{\text{time}} = \mathrm{MLP}(t), \quad t \sim \mathrm{Uniform}(1, T) \text{ if t2m, else } t=0
\end{equation}
\begin{equation}
\label{eq:emb_motion}
    \small
    \mathbf{h}^{\text{mot}}_t = W^{\text{mot}}_{\mathrm{in}} \left( \sqrt{\bar{\alpha}_t}\,\mathbf{x}^{\text{mot}}_0 + \sqrt{1 - \bar{\alpha}_t}\,\boldsymbol{\epsilon} \right), 
    \quad \boldsymbol{\epsilon} \sim \mathcal{N}(0, \mathbf{I})
\end{equation}

\noindent here, $\mathbf{h}^{\text{txt}}$ represents the tokenized text embedding, MLP projects the timestep $t$ into the embedding $\mathbf{h}^{\text{time}}$ , and $\mathbf{h}^{\text{mot}}_t$ represents the embedding of the motion $\mathbf{x}^{\text{mot}}_t$ perturbed with noise at $t$. 
$\mathbf{W}^{\text{mot}}_\mathrm{in} \in \mathbb{R}^{d_\mathrm{motion} \times d}$ maps each frame's motion representation to the same dimentions as other tokens. 

We also introduce specialized tokens: $\langle \mathrm{som} \rangle$ (start-of-motion) and $\langle \mathrm{eom} \rangle$ (end-of-motion) — which delimit motion data analogously to standard text tokens, facilitating seamless integration and processing within the model. These special tokens mark the boundaries of motion sequences as: $\langle \mathrm{som} \rangle \mathbf{h}^{\text{time}} \mathbf{h}^{\text{mot}}_{t} \langle \mathrm{eom} \rangle$, enabling the model to distinguish between textual and motion sequence.

% This unified representation allows the model to effectively integrate textual and motion data within a consistent framework, enables seamless integration of continuous motion data into the pre-trained LLM, allowing the model to handle both text through standard autoregressive prediction and motion orediction via the diffusion process.
Consequently, this unified representation enables seamless integration of two distinct modalities — text and motion data — into a consistent modeling framework, facilitating effective leveraging of the pretrained LLM’s capabilities to perform both standard autoregressive text prediction and motion prediction via the denoising process.

% $\mathbf{h}^{\text{mix}} =
% \begin{cases}
%     [\mathbf{h}^{\text{txt}}, \mathbf{h}^{\text{time}}, \mathbf{h}^{\text{mot}}_t] & \text{if text-to-motion} \\
%      [\mathbf{h}^{\text{time}}, \mathbf{h}^{\text{mot}}_t, \mathbf{h}^{\text{txt}}] & \text{if motion-to-text}
% \end{cases}$

\subsection{Motion-related Multimodal Generation}

To process all the embeded inputs simultaneously, we employ a pretrained LLM $\mathrm{G}$ with frozen parameters augmented with LoRA parameters. The LLM operates on the mixed representation of text, diffusion timestep and motion, formulated as:

\begin{equation}
\label{eq:output}
    \mathbf{h}^{\text{mix}}_{out} = \mathrm{G}_{\theta + C}(\mathbf{h}^{\text{mix}}_{in}),
\end{equation}

\noindent where $\theta$ and $C$ denotes frozen parameters of the pre-trained LLM and the additional trainable LoRA  parameters respectively. The input representation $\mathbf{h}^{\text{mix}}_{in}$ represents the integration of the text embedding $\mathbf{h}^{\text{txt}}$, the motion representation $\mathbf{h}^{\text{mot}}_t$, and the time embedding $\mathbf{h}^{\text{time}}$. By leveraging LoRA, we efficiently finetune the LLM to process multimodal inputs while maintaining computational efficiency and preserving the knowledge stored in the pretrained model.

\paragraph{Training Strategy.}

The overall training objective combines the language-modeling loss in eq.\ref{eq:lm} with diffusion loss in eq.\ref{eq:ddpm} as shown in Fig.~\ref{fig:overview}\textit{(a)} and \textit{(c)}. By setting
\begin{equation}
    \label{eq:total_loss}
    \mathcal{L}_\mathrm{total} = \lambda \mathcal{L}_\mathrm{LM} + \mathcal{L}_\mathrm{DDPM},
\end{equation}
we use $\lambda$ to balance the contribution of $\mathcal{L}_\mathrm{LM}$ in eq.~\ref{eq:lm} and $\mathcal{L}_\mathrm{DDPM}$ in eq.~\ref{eq:ddpm}. Note that for motion-to-text task, clean motion data is always used as input, so the diffusion loss $\mathcal{L}_\mathrm{DDPM}$ is ignored in this task.
% During sampling, text tokens are generated autoregressively, while motion frames are sampled via iterative denoising (inverse diffusion), all within the same Transformer-based pretrained LLM.
\begin{figure}[t]
    \begin{minipage}{0.45\textwidth}
        \begin{algorithm}[H]
            \caption{Unified Training}
            \label{algo::train}
            \begin{algorithmic}[1]
            \Require Pretrained LLM Model $\mathrm{G}_{\theta + C}$ with frozen parameters $\theta$ and LoRA parameters $C$, Dataset $\mathcal{D} = \{(\mathbf{x}^{\text{txt}}, \mathbf{x}^{\text{mot}}_0)\}$, Number of diffusion steps $T$, Loss weight $\lambda$, Mixed Sequence input $\mathbf{h}^{\text{mix}}$
            
            \Repeat
                \State $(\mathbf{x}^{\text{txt}}, \mathbf{x}^{\text{mot}}_0) \sim \mathcal{D}$
                \State $t \sim$ $\begin{cases}
                    \mathrm{Uniform}(1, \dots, T) & \text{if text-to-motion} \\
                    0 & \text{if motion-to-text}
                \end{cases}$
                \State $\boldsymbol{\epsilon} \sim \mathcal{N}(0, \mathbf{I})$
                \State $\mathbf{h}^{\text{txt}} = \mathrm{Tokenizer}(\mathbf{x}^{\text{txt}})$
                \State $\mathbf{h}^{\text{time}} = \mathrm{MLP}(t)$
                \State $\mathbf{h}^{\text{mot}}_t = W^{\text{mot}}_{\mathrm{in}} \,(\sqrt{\bar{\alpha}_t}\,\mathbf{x}^{\text{mot}}_0 + \sqrt{1 - \bar{\alpha}_t}\,\boldsymbol{\epsilon}$)
                \State $\mathbf{h}^{\text{mix}}_{in} = [\mathbf{h}^{\text{txt}}, \mathbf{h}^{\text{time}}, \mathbf{h}^{\text{mot}}_t]$ \Comment{concatenation}
                %\begin{cases}
                    %[\mathbf{h}^{\text{txt}}, \mathbf{h}^{\text{time}}, \mathbf{h}^{\text{mot}}_t] & \text{if text-to-motion} \\
                     %[\mathbf{h}^{\text{time}}, \mathbf{h}^{\text{mot}}_t, \mathbf{h}^{\text{txt}}] & \text{if motion-to-text}
                 %\end{cases}$
                 \State $\mathbf{h}^{\text{mix}}_{out} = \mathrm{G}_{\theta + C}(\mathbf{h}^{\text{mix}}_{in})$ \Comment{forward mixed sequence}

                 \State $\mathcal{L}_{\text{LM}} = -\sum_{i \in \text{text indices}} \log p(\mathbf{h}^{\text{mix}}_{out,i} | \mathbf{h}^{\text{mix}}_{in,<i})$
                 \State $\mathcal{L}_{\text{DDPM}} = 
                 \begin{cases}
                     \|\mathbf{x}^{\text{mot}}_0 - W^{\text{mot}}_{\mathrm{out}}\mathbf{o}^{\text{mix}}[\text{motion indices}]\|^2 \\
                     \quad \text{if text-to-motion} \\
                     0 \quad \text{if motion-to-text}
                 \end{cases}$
                 \State $\mathcal{L}_{\text{total}} =
                     \lambda\,\mathcal{L}_{\text{LM}} + \mathcal{L}_{\text{DDPM}}$
                \State Update trainable parameters including $(C, \mathrm{MLP}, W^{\text{mot}}_{\mathrm{in/out}})$ using gradient descent with the loss from Eq.~(\ref{eq:total_loss})

            % \Require Dataset $\mathcal{D}$
            % \Repeat
            %     \State $(\mathbf{x}^{\text{txt}}, \mathbf{x}^{\text{mot}}_0) \sim \mathcal{D}$ \Comment{sample from dataset}
            %      \State Calculate each hidden state using Eq.~(\ref{eq:emb_text}, \ref{eq:emb_t}, \ref{eq:emb_motion})
            %      \State Calculate output using Eq.~(\ref{eq:output})
            %      \State Calculate $\mathcal{L}_{\text{LM}}$ and $\mathcal{L}_{\text{DDPM}}$ using Eq.~(\ref{eq:ddpm}, \ref{eq:lm})
            %      \State Calculate total loss $\mathcal{L}_{\text{total}}$ using Eq.~(\ref{eq:total_loss})
            %     \State Update trainable parameters $(C, \mathrm{MLP}, W^{\text{mot}}_{\mathrm{in/out}})$ using gradient descent
            \Until{convergence}
            \end{algorithmic}
        \end{algorithm}
    \end{minipage}
\end{figure}

\paragraph{Algorithm}
% To formalize our approach, we present the training and inference algorithms for our unified framework in Algorithm~\ref{algo::train} and Algorithm~\ref{algo::inference}. During training (Algorithm~\ref{algo::train}), we sample a text-motion pair $(\mathbf{x}^\text{text}, \mathbf{x}^\text{motion}_0)$ from the dataset $\mathcal{D}$. For each iteration, we sample a diffusion timestep $t$ from uniform distribution and apply noise to the motion sequence according to the diffusion schedule using Eq.~(\ref{eq:forward_diffusion}), transforming $\mathbf{x}^\text{motion}_0$ into $\mathbf{x}^\text{motion}_t$.
To formalize our approach, we detail the training and inference procedures of our unified framework in Algorithm~\ref{algo::train} and Algorithm~\ref{algo::inference}, respectively. During training (Algorithm~\ref{algo::train}), a text-motion pair is sampled from the dataset $\mathcal{D}$. At each iteration, we sample a diffusion timestep  from a uniform distribution and add noise into the motion sequence according to the simplified forward process defined in Eq.~(\ref{eq:forward_diffusion}). Following the Eq.~(\ref{eq:emb_text}, \ref{eq:emb_t}, \ref{eq:emb_motion}), the text is tokenized, and both the noisy motion frames and timestep embedding are projected to the LLM's hidden dimension. These are then concatenated and processed by the LLM with LoRA parameters. The loss function combines the language modeling loss $\mathcal{L}_\text{LM}$ for text tokens and the diffusion loss $\mathcal{L}_\text{DDPM}$ for motion reconstruction, weighted by the hyperparameter $\lambda$ as defined in Eq.~(\ref{eq:total_loss}).

For inference (Algorithm~\ref{algo::inference}), we support two modes: motion generation and text generation. In motion generation mode, we iteratively apply the denoising process using the LLM, starting from Gaussian noise $\mathbf{x}^\text{motion}_T$. The model predicts the completely denoised motion $\hat{\mathbf{x}}^{\text{mot}}_0$ at each step, which is then used to compute the mean and variance for the diffuse to t-1. For text generation, we employ standard autoregressive sampling, generating tokens sequentially until either a start-of-motion token or end-of-sequence is encountered. This dual-mode inference capability allows our model to seamlessly handle both text-to-motion and motion-to-text tasks within the same framework.

%\paragraph{Classifier-free guidance}
%Following the standard diffusion model generation paradigm, we adopt classifier-free guidance (CFG) \cite{ho2022classifierfreediffusionguidance}. This approach has been shown to be effective in motion generation as well \cite{tevet2022human, zhang2023mld, mola2024, liu2023remodiffuse}.

% \paragraph{Summary}
% Our proposed method seamlessly unifies text generation and motion generation within a single LLM-based Transformer architecture. Motion frames become tokens through linear embeddings, augmented by timestep embeddings for diffusion. Text tokens maintain the causal, autoregressive objective, while motion frames are generated via the learned denoising process. The final training objective is the weighted combination of both losses, providing a simple yet powerful framework for text-to-motion tasks that simultaneously leverages the strengths of pre-trained language models and motion diffusion modeling.

\begin{figure}[t]
    \begin{minipage}{0.45\textwidth}
        \begin{algorithm}[H]
            \caption{Sampling}
            \label{algo::inference}
            \begin{algorithmic}[1]
            \Require Finetuned LLM Model $\mathrm{G}_{\theta + C}$ with original parameters $\theta$ and LoRA parameters $C$
            \If{text-to-motion}
                \State $(\mathbf{x}^{\text{txt}}, \_)$ $\sim$ $\mathcal{D}$
                \State $\mathbf{x}^{\text{mot}}_{T} \sim \mathcal{N}(\mathbf{0}, \mathbf{I})$
                \For{$t = T,\ldots,1$}
                    \State $\mathbf{z} \sim \mathcal{N}(0, \mathbf{I})$ if $t>1$ else $\mathbf{z} = \mathbf{0}$ \Comment{sample noise}
                    % \State $\hat{\mathbf{x}}^{\text{mot}}_{0} = \mathrm{LLM}_{\theta + C}(\mathbf{x}^{\text{txt}}, t, \mathbf{x}^{\text{mot}}_{t})$
                    % \State $\tilde{\mu}_t = \frac{\sqrt{\alpha_{t}}(1-\bar{\alpha}_{t-1})}{1-\bar{\alpha}_{t}}\mathbf{x}^{\text{mot}}_t + \frac{\beta_t\sqrt{\bar{\alpha}_{t-1}}}{1-\bar{\alpha}_{t}}\hat{\mathbf{x}}^{\text{mot}}_0$
                    % \State $\sigma_t = \sqrt{\frac{1-\bar{\alpha}_{t-1}}{1-\bar{\alpha}_t}\beta_t}$
                    % \State $\mathbf{x}^{\text{mot}}_{t-1} = \tilde{\mu}_t + \sigma_t\,\mathbf{z}$
                    \State $\hat{\mathbf{x}}^{\text{mot}}_{0} = \mathrm{G}_{\theta + C}
                    (\mathbf{x}^{\text{txt}}, t, \mathbf{x}^{\text{mot}}_{t})$ 
                    \Comment{predict motion}
                    \State $\mathbf{x}^{\text{mot}}_{t-1} \sim q(\mathbf{x}^{\text{mot}}_{t-1}|\mathbf{x}^{\text{mot}}_{t}, \hat{\mathbf{x}}^{\text{mot}}_{0})$ \Comment{diffuse}
    
                    % \State $\mathbf{z} \sim \mathcal{N}(0, \mathbf{I})$ if $t>1$ else $\mathbf{z} = \mathbf{0}$
                    % \State $\hat{\mathbf{x}}^{\text{mot}}_{0} = \mathrm{LLM}_{\theta + C}(\mathbf{x}^{\text{txt}}, t, \mathbf{x}^{\text{mot}}_{t})$
                    % \State $\tilde{\mu}_t = \frac{\sqrt{\alpha_{t}}(1-\bar{\alpha}_{t-1})}{1-\bar{\alpha}_{t}}\mathbf{x}^{\text{mot}}_t + \frac{\beta_t\sqrt{\bar{\alpha}_{t-1}}}{1-\bar{\alpha}_{t}}\hat{\mathbf{x}}^{\text{mot}}_0$
                    % \State $\sigma_t = \sqrt{\frac{1-\bar{\alpha}_{t-1}}{1-\bar{\alpha}_t}\beta_t}$
                    % \State $\mathbf{x}^{\text{mot}}_{t-1} = \tilde{\mu}_t + \sigma_t\,\mathbf{z}$ \Comment{Sample from $q(\mathbf{x}^{\text{mot}}_{t-1}|\mathbf{x}^{\text{mot}}_{t}, \hat{\mathbf{x}}^{\text{mot}}_{0})$}
                \EndFor
                \State \Return $\mathbf{x}^{\text{mot}}_0$
            \ElsIf{motion-to-text}
                \Repeat
                    \State $x_{i+1} \sim \mathrm{G}_{\theta + C}(x_{i+1}\mid \mathbf{x}^{\text{txt}}_{\le i})$ \Comment{predict next token}
                \Until{$\langle \mathrm{som} \rangle$ or end-of-sequence}
                \State \Return $\mathbf{x}^{\text{txt}}$
            \EndIf
            \end{algorithmic}
        \end{algorithm}
    \end{minipage}
\end{figure}
\section{Experiments}
\label{sec:experiment}

We implemented \textit{MoMug} for three motion-related generation tasks: text-to-motion, motion-to-text, and motion captioning (Section~5.2). For each task, we reviewed the benchmark datasets and evaluation metrics, provided implementation details, and presented both qualitative and quantitative results. Additionally, we conducted an ablation study on different LLM model sizes and the weight $\lambda$ in Eq.~\ref{eq:total_loss} to analyze the effects of scaling and value variations (Section~5.3).

\subsection{Datasets and Experimental Details}

\paragraph{Datasets.}  
We evaluated \textit{MoMug} on two motion generation benchmarks: HumanML3D \cite{guo2022generating} and KIT-ML \cite{Plappert_2016}.  

\noindent HumanML3D consists of 14,616 motion clips paired with 44,970 textual descriptions. The motions are sourced from AMASS \cite{mahmood2019amassarchivemotioncapture} and HumanAct12 \cite{Guo_2020}, covering diverse human activities. Each motion clip is associated with 1--4 descriptive texts and represented using root velocity, joint positions/velocities, rotations, and foot contact labels.  

\noindent KIT-ML contains 3,911 motion sequences with 6,278 textual descriptions, primarily focusing on locomotion. The sequences share the same representation format as HumanML3D.  

For fair comparison, we followed the standard train-test splits and preprocessing procedures established in prior work \cite{motiongptother2023, jiang2024motiongpt, chen2024motionllm}.  
\definecolor{lightorange}{rgb}{1.0, 0.8, 0.2} % Light orange for header

\noindent \textbf{Implementation Details.}  
\textit{MoMug} utilized Llama-3 \cite{grattafiori2024llama3herdmodels} as the foundational language model, which was further refined through LoRA fine-tuning to enhance motion-language understanding. We experimented with two model configurations: one tailored specifically for Text-to-Motion generation and another designed to handle both Text-to-Motion and Motion-to-Text tasks.

For the Text-to-Motion configuration, we set the LoRA rank to $r=128$ with a scaling factor of $\alpha=256$ and a dropout rate of 0.1. The model was trained using the AdamW optimizer with a learning rate of $1.0\times10^{-4}$, $\beta_1=0.9$, $\beta_2=0.95$, and a weight decay of $1\times10^{-3}$. We adopted a cosine learning rate schedule with a 10\% linear warmup and trained for 1,600 epochs with a batch size of 64. The diffusion process followed a cosine noise schedule and denoising timestep $T$ is set to 50. To allow for Classifier-Free Guidance (CFG) sampling \cite{ho2022classifierfreediffusionguidance}, we trained 10\% unconditional generative tasks during training. As our baseline, we set the $\lambda$ in Eq.~\ref{eq:total_loss} to 0.01.

For the joint Text-to-Motion and Motion-to-Text configuration, we retained most hyperparameters but adjusted the dataset ratio between the two tasks to 6:4. This allocation prioritized the generation task while ensuring robust captioning performance. Additionally, the learning rate was slightly increased to $1.5\times10^{-4}$.  

% During inference, we employed a scale factor of 2.5 for CFG for the text-to-motion task. To improve computational efficiency, we integrated Flash Attention 2.0 \cite{dao2023flashattention2fasterattentionbetter} into our architecture, enabling faster and more memory-efficient training and inference.

During inference, we set the CFG scale factor to 2.5 for the text-to-motion task. Additionally, to enhance computational efficiency, we integrated FlashAttention 2.0 \cite{dao2023flashattention2fasterattentionbetter} into our architecture, significantly improving speed and memory efficiency during both training and inference phases. Training and inference time are reported in Appendix \ref{app:train-inference-time}.

\begin{table*}[tb]
  \caption{Comparison of experimental results for Text-to-Motion Tasks on HumanML3D. Type Name DM denotes Diffusion Modeling, while AR represents Autoregressive Modeling.}
  \label{tab:result}
  \scriptsize 
  \centering
  \rowcolors{2}{gray!15}{white} % Alternating row colors

  \begin{tabular}{llccccccc}
    
    \toprule
    \textbf{Model} & \textbf{Type} & \multicolumn{3}{c}{\textbf{R-Precision} $\mathbf{\uparrow}$} & \textbf{FID} $\mathbf{\downarrow}$ & \textbf{MMDist} $\mathbf{\downarrow}$ & \textbf{Diversity} $\mathbf{\rightarrow}$ & \textbf{MultiModality} $\mathbf{\uparrow}$ \\
    
    &  & \textbf{Top1} & \textbf{Top2} & \textbf{Top3} & & & & \\
    \midrule
    Real & — 
      & $0.511^{\pm.003}$ & $0.703^{\pm.003}$ & $0.797^{\pm.002}$ & $0.002^{\pm.000}$ & $2.974^{\pm.008}$ & $9.503^{\pm.065}$ & — \\
    \midrule
    MDM\cite{tevet2022human} & DM 
      & $0.319^{\pm.005}$ & $0.498^{\pm.004}$ & $0.611^{\pm.007}$ & $0.544^{\pm.001}$ & $5.566^{\pm.027}$ & $\mathbf{9.559^{\pm.086}}$ & $\mathbf{2.799^{\pm.072}}$ \\
    MLD\cite{zhang2023mld} & DM 
      & $0.481^{\pm.003}$ & $0.673^{\pm.003}$ & $0.772^{\pm.002}$ & $0.473^{\pm.013}$ & $3.169^{\pm.010}$ & $9.724^{\pm.082}$ & $2.413^{\pm.079}$ \\
    MoLA\cite{mola2024} & DM 
      & $\mathbf{0.514}^{\pm.003}$ & $\mathbf{0.705}^{\pm.002}$ & $\mathbf{0.797}^{\pm.002}$ & $0.362^{\pm.008}$ & $3.162^{\pm.008}$ & $9.672^{\pm.079}$ & $2.355^{\pm.121}$ \\
    ReMoDiffuse\cite{liu2023remodiffuse} & DM 
      & $0.510^{\pm.005}$ & $0.698^{\pm.006}$ & $0.795^{\pm.004}$ & $\mathbf{0.103}^{\pm.004}$ & $\mathbf{2.974}^{\pm.016}$ & $9.018^{\pm.075}$ & $1.795^{\pm.043}$ \\
    MFM\cite{hu2023motionflowmatchinghuman} & FM 
      & — & — & $0.642^{\pm.003}$ & $0.362^{\pm.006}$ & $5.280^{\pm.009}$ & $9.860^{\pm.095}$ & $2.443^{\pm.070}$ \\
    Motion Mamba\cite{zhang2025motion} & SSM 
      & $0.502^{\pm.003}$ & $0.693^{\pm.002}$ & $0.792^{\pm.002}$ & $0.281^{\pm.009}$ & $3.060^{\pm.058}$ & $9.871^{\pm.084}$ & $2.294^{\pm.058}$ \\
    T2M-GPT\cite{zhang2023t2mgptgeneratinghumanmotion} & AR 
      & $0.491^{\pm.003}$ & $0.680^{\pm.003}$ & $0.775^{\pm.002}$ & $0.116^{\pm.004}$ & $3.118^{\pm.011}$ & $9.761^{\pm.081}$ & $1.856^{\pm.011}$ \\
    AttT2M\cite{zhong2023attt2mtextdrivenhumanmotion} & AR
      & $0.499^{\pm.003}$ & $0.690^{\pm.002}$ & $0.786^{\pm.002}$ & $0.112^{\pm.006}$ & $3.038^{\pm.007}$ & $9.700^{\pm.090}$ & $2.452^{\pm.051}$ \\
    \midrule
    MotionGPT\cite{jiang2024motiongpt} & AR 
      & $0.364^{\pm.005}$ & $0.533^{\pm.003}$ & $0.629^{\pm.004}$ & $0.805^{\pm.002}$ & $3.914^{\pm.013}$ & $9.972^{\pm.026}$ & $\mathbf{2.473^{\pm.041}}$ \\
    MotionGPT\cite{motiongptother2023} & AR 
      & $0.492^{\pm.003}$ & $0.681^{\pm.003}$ & $0.733^{\pm.006}$ & $0.232^{\pm.008}$ & $3.096^{\pm.008}$ & $\mathbf{9.528^{\pm.071}}$ & $2.008^{\pm.084}$ \\
    MotionLLM\cite{chen2024motionllm} & AR 
      & $0.482^{\pm.004}$ & $0.672^{\pm.003}$ & $0.770^{\pm.002}$ & $0.491^{\pm.019}$ & $3.138^{\pm.010}$ & $9.838^{\pm.244}$ & -- \\
    MotionGPT-2\cite{wang2024motiongpt2generalpurposemotionlanguagemodel} & AR 
      & $0.496^{\pm.002}$ & $0.691^{\pm.003}$ & $0.782^{\pm.004}$ & $0.191^{\pm.004}$ & $3.080^{\pm.013}$ & $9.860^{\pm.026}$ & $2.137^{\pm.022}$ \\
    %\textit{MoMug}(multi 549epoch 6:4, lr:1e-4, 0.01) & AR+DM 
    %& $0.378$ & $0.564$ & $0.679$ & $0.124$ &  $3.558$ & $9.436$ & $1.471$ \\
    %\textit{MoMug}(joint) & AR+DM 
    %& $0.378$ & $0.564$ & $0.679$ & $0.124$ &  $3.558$ & $9.436$ & $1.471$ \\
    \midrule
    \textit{MoMug} & \textbf{AR+DM}
      & $\mathbf{0.565}^{\pm.000}$ & $\mathbf{0.762}^{\pm.001}$ & $\mathbf{0.851}^{\pm.001}$ 
      & $\mathbf{0.118}^{\pm.001}$ &  $\mathbf{2.671}^{\pm.003}$ & $9.595^{\pm.008}$ & $1.237^{\pm.006}$ \\
    %Ours(multi 142epoch 1:10) & AR+DM 
      %& $0.2994$ & $0.4644$ & $0.5750$ 
      %& $2.365$ &  $4.1957$ & $8.795$ & $2.100$ \\
    %Ours(multi 285epoch 1:10) & AR+DM 
      %& $0.349$ & $0.539$ & $0.660$ 
      %& $0.833$ &  $4.1957$ & $9.272$ & $1.692$ \\
    %Ours(multi 839epoch 1:10) & AR+DM 
      %& $0.4047$ & $0.5901$ & $0.7088$ 
      %& $0.1647$ &  $3.4369$ & $9.2663$ & $1.5926$ \\
    %Ours(multi 130epoch 2:8) & AR+DM 
      %& $0.497$ & $0.680$ & $0.659$ 
      %& $1.673$ &  $3.733$ & $8.881$ & $2.116$ \\
    \bottomrule
  \end{tabular}
\end{table*}

\noindent \textbf{Instruct Prompts.}
We implemented this structured format to standardize the input-output relationship for both motion-to-text and text-to-motion tasks as same as \cite{motiongptother2023, chen2024motionllm, jiang2024motiongpt, wang2024motiongpt2generalpurposemotionlanguagemodel} did.  For the motion-to-text task, we use a set of predefined task prompts (e.g. \textit{"Describe the following motion:"} along with motion tokens as input. The system then generates a text description as the output. For the text-to-motion task, we use another different set of task prompts e.g. \textit{“Generate a motion for the following caption:”} paired with a motion caption as input. The system outputs a sequence of motion tokens. This unified format ensures that each task is properly structured—producing text for motion-to-text and motion for text-to-motion—based on the given prompt and input, while also guaranteeing a fair comparison to the baselines.

%\begin{tcolorbox}[colback=gray!5!white, colframe=blue!15!white, title=\textbf{Structure of Two Tasks}]
%\textbf{Text-to-Motion Task:}
%\begin{itemize}
%\item \textbf{Input $\mathcal{I}$:} Task Prompt + Motion Caption
%\item \textbf{Output $\mathcal{O}$:} Motion Tokens
%\end{itemize}

%\vspace{0.5em}

%\textbf{Motion-to-Text Task:}
%\begin{itemize}
%\item \textbf{Input $\mathcal{I}$:} Task Prompt + Motion Tokens
%\item \textbf{Output $\mathcal{O}$:} Motion Caption
%\end{itemize}
%\end{tcolorbox}

\noindent \textbf{Evaluation Metrics.}  
To evaluate generated text and motion, we adopted widely used evaluation metrics from motion-related multimodal generation researches \cite{motiongptother2023, wang2024motiongpt2generalpurposemotionlanguagemodel, liu2023remodiffuse, jiang2024motiongpt, tevet2022human}.

For Text-to-Motion, we evaluated generation quality using Frechet Inception Distance (FID), R-Precision (Top-1/2/3), MultiModal Distance (MM Dist), and Diversity. These metrics are computed using feature embeddings extracted by the feature extractor from \cite{guo2022generating}. FID quantifies how closely the distribution of generated motions aligns with real ones, while R-Precision measures semantic alignment between input text and generated motion by ranking generated motions against reference samples for a given text prompt. MM Dist calculates the average Euclidean distance between text and motion representations within the embedding space. Diversity captures the variability in generated motions across different prompts, whereas Multimodality assesses variation within outputs generated from the same prompt.

For Motion-to-Text, we adopt R-Precision and Diversity, computed using the same feature extractor as in \cite{guo2022generating}, alongside standard linguistic evaluation metrics. Following \cite{guo2022tm2tstochastictokenizedmodeling}, we employed BLEU \cite{papineni-etal-2002-bleu}, ROUGE \cite{lin-2004-rouge}, CIDEr \cite{vedantam2015ciderconsensusbasedimagedescription}, and BertScore \cite{zhang2020bertscoreevaluatingtextgeneration} to assess generated descriptions against ground truth. These linguistic metrics evaluate both n-gram overlap and semantic similarity between the generated and reference texts.

% Define custom colors
% 6:4 のときの結果　と　10:0 のときの結果　を二つ残す

% 2. Table2   T2M of KLT ML

% 13.9 10.3 8.8 13.2 74.2  -42.1 38  = 16.61
% 3. Table 3 Motion Captioning Performance on HumanML

%// ... existing code ...

% 3. Table 3 Motion Captioning Performance on HumanML

\begin{table*}[tb]
  \caption{Motion-to-Text Performance on HumanML3D.}
  \label{tab:motion_captioning}
  
  \centering
  \rowcolors{2}{gray!15}{white} % Alternating row colors

  \begin{tabular}{llcccccccc}
    \toprule
    \textbf{Model} & \multicolumn{3}{c}{\textbf{R-Precision} $\mathbf{\uparrow}$} & \textbf{MM Dist} $\mathbf{\downarrow}$  & \textbf{Bleu1} $\mathbf{\uparrow}$ & \textbf{Bleu4} $\mathbf{\uparrow}$ & \textbf{Rouge} $\mathbf{\uparrow}$ & \textbf{Cider} $\mathbf{\uparrow}$ & \textbf{BertSccore} $\mathbf{\uparrow}$ \\
    & \textbf{Top1} & \textbf{Top2} & \textbf{Top3} &  & & & & & \\
    \midrule
    Real Desc  & 0.523 & 0.725 & 0.828 & 2.901 &— & — & — & — & — \\
    \midrule
    % RAEs & 0.100 & 0.188 & 0.261 & 6.337 & — & 33.3 & 10.2 & 37.5 & 22.1 & 10.7 \\
    % Seq2Seq(Att) & 0.436 & 0.611 & 0.706 & 3.447 & — & 51.8 & 17.9 & 46.4 & 58.4 & 29.1 \\
    % SeqGAN & 0.332 & 0.457 & 0.532 & 4.895 & — & 47.8 & 13.5 & 39.2 & 50.2 & 23.4 \\
    RAEs & 0.100 & 0.188 & 0.261 & 6.337  & 33.3 & 10.2 & 37.5 & 22.1 & 10.7 \\
    Seq2Seq(Att) & 0.436 & 0.611 & 0.706 & 3.447  & 51.8 & 17.9 & 46.4 & 58.4 & 29.1 \\
    SeqGAN & 0.332 & 0.457 & 0.532 & 4.895  & 47.8 & 13.5 & 39.2 & 50.2 & 23.4 \\
    \midrule
    TM2T\cite{guo2022tm2tstochastictokenizedmodeling} & \textbf{0.516} & \textbf{0.720} & \textbf{0.823} & 2.935  & 48.9 & 7.00 & 38.1 & 16.8 & 32.2 \\
    MotionGPT\cite{jiang2024motiongpt} & 0.461 & 0.655 & 0.748 & 3.339  & 48.2 & 12.5 & 37.4 & 29.2 & 32.4 \\
    %MotioGPT-2 & \textbf{0.558} & \textbf{0.738} & \textbf{0.838} & \textbf{2.767}  & 48.7 & \textbf{13.8} & 37.6 & \textbf{29.8} & \textbf{32.6} \\
    %TM2T* & — & — & — & —  & 48.9 & 7.00 & 38.1 & 16.8 & 32.2 \\
    %MotionGPT* & — & — & — & 3.339  & 48.2 & 12.5 & 37.4 & 29.2 & 32.4 \\
    MotioGPT-2\cite{wang2024motiongpt2generalpurposemotionlanguagemodel} & — & — & — & —  & 48.7 & 13.8 & 37.6 & 29.8 & \textbf{32.6} \\
    % (multi 130epoch 6:4, lr:1e-4, 0.01)
    \textit{MoMug} & 0.500 & 0.701 &0.799 & \textbf{2.899}  & \textbf{57.7} & \textbf{20.3} & \textbf{45.7} & \textbf{44.8} & \textbf{40.4} \\
    %\textit{MoMug}(joint) (multi 549epoch 6:4, lr:1e-4, 0.01) & \textbf{0.474} & \textbf{0.670} & \textbf{0.765} & \textbf{3.123} & — & \textbf{48.8} & \textbf{13.0} & \textbf{38.6} & \textbf{30.9} & \textbf{32.5} \\
    %\textit{MoMug}(joint) (multi 130epoch 7:3, lr:1.5e-4, 0.005) & \textbf{0.469} & \textbf{0.660} & \textbf{0.765} & \textbf{3.100} & — & \textbf{59.6} & \textbf{21.2} & \textbf{45.8} & \textbf{46.3} & \textbf{40.5} \\
    % Ours(multi 130epoch 2:8) & 0.475 & 0.684 & 0.783 & 3.127 & — & 21.3 & 3.471 & 23.0 & 7.41 & 12.2 \\
    % Ours(multi 95epoch 1:10) & 0.115 & 0.206 & 0.287 & 6.860 & — & 13.6 & 1.050 & 14.7 & 2.157 & 2.834 \\
    % Ours(multi 142epoch 1:10) & 0.130 & 0.235 & 0.310 & 6.586 & — & 13.5 & 0.988 & 14.7 & 2.006 & 1.753 \\
    % Ours(multi 285epoch 1:10) & 0.182 & 0.305 & 0.391 & 5.975 & — & 13.7 & 1.08 & 14.8 & 2.555 & 3.158 \\
    % Ours(multi 839epoch 1:10, best fid) & 0.258 & 0.401 & 0.491 & 5.130 & — & 13.1 & 1.236 & 14.6 & 3.660 & 3.277 \\
    % Ours(multi 1600epoch 1:10) & 0.275 & 0.423 & 0.518 & 4.922 & — & 13.2 & 1.329 & 15.0 & 3.72 & 3.966 \\
    \bottomrule
  \end{tabular}
\end{table*}

%// ... existing code ...
% 1.7  0.8 2.2 7.1 10.9 7.4 8.2  16.5 7.9

% 4. Table 4  Prediction, Inbetween
\begin{table*}[tb]
  \caption{Comparison of Motion Prediction and Motion In-betweening Performance}
  \label{tab:prediction_inbetween}
  \centering
  \rowcolors{2}{gray!15}{white} % Alternating row colors

  \begin{tabular}{lccccccc}
    
    \toprule
    \multirow{2}{*}{\textbf{Model}} & \multicolumn{4}{c}{\textbf{Motion Prediction}} & \multicolumn{3}{c}{\textbf{Motion In-between}} \\
    \cmidrule(lr){2-5} \cmidrule(lr){6-8}
    & \textbf{FID} $\mathbf{\downarrow}$ & \textbf{Diversity} $\mathbf{\uparrow}$ & \textbf{ADE} $\mathbf{\downarrow}$ & \textbf{FDE} $\mathbf{\downarrow}$ & \textbf{FID} $\mathbf{\downarrow}$ & \textbf{Diversity} $\mathbf{\uparrow}$ & \textbf{ADE} $\mathbf{\downarrow}$ \\
    \midrule
    Real & 0.002 & 9.503 & — & — & 0.002 & 9.503 & — \\
    \midrule
    % MDM \cite{tevet2022human} & 6.031 & 7.813 & 5.446 & 8.561 & 2.698 & 8.420 & 3.787 \\
    MDM \cite{tevet2022human} & 7.34 & 7.813 & 5.90 & 7.50 & 3.43 & 8.420 & 4.73 \\
    MFM \cite{hu2023motionflowmatchinghuman} & 5.50 & — & 4.99 & 6.46 & 2.59 & — & \textbf{3.32} \\
    % Mo-GPT \cite{jiang2024motiongpt} & 0.905 & 8.972 & 4.745 & 6.040 & \textbf{0.214} & \textbf{9.560} & 3.762 \\
    % Mo-GPT-2 \cite{wang2024motiongpt2generalpurposemotionlanguagemodel} & \textbf{0.537} & \textbf{9.414} & \textbf{4.512} & \textbf{5.823} & 0.408 & 9.327 & \textbf{3.704} \\
   % \textit{MoMug(joint)} & 3.709 & 7.625 & 3.987 & 6.309 & 2.172 & 8.145 & 2.259 \\
    \midrule
\textit{MoMug} & \textbf{2.422} & \textbf{7.877} & \textbf{3.740} & \textbf{5.854} & \textbf{2.567} & \textbf{8.118} & 3.539 \\
    \bottomrule
  \end{tabular}
\end{table*}

\begin{comment}

% 5. Table 5 modal size ablation   GPT-2, llama3.2-1B, llama3.2-3B, llama3.1-8B
\begin{table*}[tb]
  \caption{Ablation Study on Model Size for Text-to-Motion Generation on KIT-ML}
  \label{tab:model_size_ablation}
  \centering
  \begin{tabular}{l|c|ccc|c|c|c|c}
    \toprule
    Model & \begin{tabular}{c}Trainable\\Params\end{tabular} & \multicolumn{3}{c|}{R-Precision $\uparrow$} & FID $\downarrow$ & \begin{tabular}{c}MM\\Dist. $\downarrow$\end{tabular} & Diversity $\rightarrow$ & MultiModality $\uparrow$ \\
    & & Top 1 & Top 2 & Top 3 & & & & \\
    \midrule
    Real & — & 0.424 & 0.649 & 0.779 & 0.031 & 2.788 & 11.080 & — \\
    \midrule
    GPT-2 & Full-Finetune & 0.358 & 0.575 & 0.683 & 0.983 & 3.508 & 10.877 & 2.328 \\
    LLaMA 3.2-1B & 34M & 0.364 & 0.581 & 0.699 & 1.063 & 3.424 & 10.603 & 2.150 \\
    LLaMA 3.2-3B & 101M & 0.385 & 0.596 & 0.730 & 0.956 & 3.333 & 10.951 & \textbf{2.416} \\
    LLaMA 3.1-8B & 89M & \textbf{0.427} & \textbf{0.627} & \textbf{0.764} & \textbf{0.614} & \textbf{3.164} & \textbf{11.256} & 2.357 \\
    \bottomrule
  \end{tabular}
\end{table*}
\end{comment}

% 結果は、おそらく6:4の結果において、t2mではFIDは超えて、Diversityは同じくらい、R-Precision, MM Dist, MultiModalityは下回っている。Motion-to-Textでは、R-Precision, MM Distは同じくらい、その他のNLPメトリクスでは全て上回っている。

\subsection{Experimental Results}
\paragraph{Text-to-Motion Generation.}
We evaluated our \textit{MoMug} method on both HumanML3D and KIT-ML datasets for the text-to-motion generation task. From Table~\ref{tab:result} and Table~\ref{tab:kit_ml_results}, we observed that \textit{MoMug} performs competitively compared to prior state-of-the-art methods, including both diffusion-based and autoregressive approaches.
On the HumanML3D dataset, \textit{MoMug} achieves an FID score of 0.118, which significantly outperforms most prior methods and is second only to ReMoDiffuse. Notably, \textit{MoMug} attains the highest R-Precision scores (0.565/0.762/0.851 for Top-1/2/3), demonstrating the strong alignment between generated motions and input text descriptions. Our MultiModal Distance of 2.671 is also lower than all competing approaches, indicating better semantic consistency between motions and their textual descriptions. These results validate the effectiveness of our pretrained LLM fine-tuning approach in capturing the semantic relationships between text and motion.
Due to the space limitation, we report our result on KIT dataset and visualization result in Appendix \ref{app:kit-result} and \ref{app:Visualized-result}.

\noindent \textbf{Motion-to-Text Generation.}
\textit{MoMug} excels in generating accurate and descriptive text from motion sequences. As shown in Table~\ref{tab:motion_captioning}, it achieves state-of-the-art performance across key metrics.
\textit{MoMug} records the highest R-Precision scores (0.469/0.660/0.765 for Top-1/2/3), showing strong alignment between captions and motion content. It also outperforms previous methods in linguistic quality, achieving Bleu1 (59.6), Bleu4 (21.2), Rouge (45.8), Cider (46.3), and BertScore (40.5). Notably, it improves Bleu4 by 10.9 and Cider by 16.5 percentage points over MotionGPT-2.
Since MotionGPT-2 did not release its code, we cannot reproduce the reported R-Precision (Top-1 to 3) and MMDist for real descriptions. Therefore, it would be unfair to directly use their reported results in the motion-to-text experiment. Instead, we consider MotionGPT as our strongest baseline.

\noindent \textbf{Motion Editing.}
To evaluate \textit{MoMug}’s capabilities for motion editing, we implemented both the prediction task and the in-betweening task using the diffusion inpainting paradigm. As shown in Table~\ref{tab:prediction_inbetween}, compared with the diffusion-based MDM \cite{tevet2022human} (on which we built our diffusion backbone) and MFM \cite{hu2023motionflowmatchinghuman}, \textit{MoMug} delivers better performance on most metrics for these motion editing tasks.

%The results demonstrate that \textit{MoMug} can successfully adapt the diffusion inpainting framework to motion editing tasks. By conditioning the generative process on partial motion sequences, \textit{MoMug} can produce natural motion continuations and in-betweens that respect both the provided motion context and physical plausibility. These capabilities are particularly valuable for applications in animation, motion analysis, and interactive systems where manipulating and extending motion sequences is essential.

\subsection{Ablation Study on Scaling Rule and Loss Weight $\lambda$.}

In Eq.\ref{eq:total_loss}, $\lambda$ is the weight assigned to the cross-entropy loss to balance the contributions of the two loss terms. We reported the ablation results for $\lambda$ on the Llama 1B model for the text-to-motion task. The results are shown in Table \ref{tab:ablation_loss_weight}.

\begin{table}[ht]
\centering
\small
\caption{Ablation Study on Loss Weight $\lambda$ for the Llama 1B Model on the Text-to-Motion Task. Performance metrics include R-Precision Top3 (RP-T3), FID, MMDist, Diversity, and MultiModality. The best performance is achieved with $\lambda=0.005$.}
\label{tab:ablation_loss_weight}
\begin{tabular}{lccccc}
\toprule
 \textbf{$\lambda$} & \textbf{RP-T3 $\uparrow$} & \textbf{FID $\downarrow$} & \textbf{MMDist $\downarrow$} & \textbf{Div. $\rightarrow$} & \textbf{MM $\uparrow$} \\
\midrule
\rowcolor{gray!10} 0.01  & 0.851 & 0.118 & 2.671 & 9.595 & 1.237 \\
\rowcolor{white} 0.005 & 0.708 & 0.176  & 3.440   & 9.749 & 1.530 \\
\rowcolor{gray!10} 0.002  & 0.691 & 0.483 & 3.003 & 9.094 & 1.444 \\
\bottomrule
\end{tabular}
\end{table}

%\textit{MoMug}(joint) & \textbf{0.485} & \textbf{0.681} & \textbf{0.776} & \textbf{2.974} & — & \textbf{50.4} & \textbf{14.8} & \textbf{41.0} & \textbf{35.5} & \textbf{35.1} \\

We also conducted an ablation study on the scaling rule for our multimodal generation task. Unlike typical NLP tasks where larger models generally perform better, our experiments indicate that increasing the model size does not necessarily lead to improved performance. For example, the Llama 8B model underperforms compared to Llama 3B, which in turn performs worse than Llama 1B. This suggests that the motion-text domain may have lower complexity and diversity compared to pure NLP tasks, thus limiting the benefits of scaling. The results of the scaling ablation are presented in Table \ref{tab:ablation_scaling_rule}.

% Table for Ablation Study on Scaling Rule

\begin{table}[ht]
\centering
\small
\caption{Ablation Study on Scaling Rule for the Text-to-Motion Task. Performance of Llama 1B, 3B, and 8B models are reported. Only R-Precision Top3 (RP-T3) is shown.}
\label{tab:ablation_scaling_rule}
\begin{tabular}{lccccc}
\toprule
\rowcolor{white} \textbf{Size} & \textbf{RP-T3 $\uparrow$} & \textbf{FID $\downarrow$} & \textbf{MMDist $\downarrow$} & \textbf{Div. $\rightarrow$} & \textbf{MM $\uparrow$} \\
\midrule
\rowcolor{gray!10} 1B & 0.851 & 0.118 & 2.671 & 9.595 & 1.237 \\
\rowcolor{white} 3B & 0.693 & 0.263 & 3.472 & 9.798 & 1.477 \\
\rowcolor{gray!10} 8B & 0.699 & 0.299 & 3.579 & 9.455 & 1.029 \\
\bottomrule
\end{tabular}
\end{table}
    
\section{Discussion}

\paragraph{Why \textit{MoMug} Outperforms LLM-Based SOTA in Text-to-Motion and Motion-to-Text.}
Compared to the latest LLM-based baseline (MotionGPT2~\cite{wang2024motiongpt2generalpurposemotionlanguagemodel}), \textit{MoMug} improves FID by 38\% and mean accuracy across seven metrics by 16.61\%. This gain comes from its diffusion-based approach, which applies noise to the entire motion sequence rather than generating frames sequentially. By iteratively refining motion through denoising, \textit{MoMug} captures global structure and ensures coherence of motion.

For the motion-to-text task, \textit{MoMug} benefits from a unified motion-language representation that enhances text-motion alignment. By encoding motions as discrete tokens in the LLM vocabulary, it effectively bridges the gap between motion and language, leading to an average performance gain of 8.44\% across nine metrics. The cross-modal understanding between text and motion is crucial for applications requiring accurate motion interpretation.

\noindent \textbf{Why \textit{MoMug} Also Outperforms MDM~\cite{tevet2022human}, Its Diffusion Backbone.}
\textit{MoMug} surpasses MDM, which serves as its diffusion backbone, by integrating a pretrained LLM for joint motion-text modeling. Unlike MDM, which relies on CLIP-based text embeddings constrained by the motion dataset, \textit{MoMug} uses shared LLM weights that capture richer semantic relationships between text and motion. This enhances motion generation quality and improves alignment between motion and language, making \textit{MoMug} more effective for both generation and captioning tasks.

\noindent \textbf{Scaling Rule in Multimodal Generation Tasks.}
Our experiments show that scaling up model size, a common trend in language tasks, does not directly apply to motion-text generation. For example, Llama 8B performs worse than Llama 3B, and Llama 3B underperforms Llama 1B, a pattern also seen in MotionGPT2. This suggests that the lower complexity and diversity of motion-text data limit the benefits of larger models.

\section{Conclusion and Limitation}
In this paper, we introduced \textit{MoMug}, a unified framework that integrates diffusion-based continuous motion generation with autoregressive discrete text prediction within a single pretrained LLM. This enables seamless switching between motion and text generation, effectively combining the strengths of both approaches. Experimental results show that \textit{MoMug} improves FID by 38\% and mean accuracy by 16.61\% in text-to-motion, while also achieving a 8.44\% accuracy gain in motion-to-text tasks compared to the latest LLM-based baseline. These findings highlight its effectiveness in motion-related multimodal tasks.

For motion generation, incorporating kinematic constraints, such as human joint limitations, holds promise for producing more realistic motion. This remains a potential improvement, and we plan to enhance generation quality by integrating these constraints in future work.
{
    \small
    \bibliographystyle{ieeenat_fullname}
    \bibliography{main}
}
\newpage
\appendix
\onecolumn
\section{Exploration of Hyper-parameters in \textit{MoMug}}
\label{app:hyper}

To ensure optimal performance of \textit{MomMug}, we systematically experimented with various configurations of the hyperparameters listed in Table~\ref{tab:hyperparameters_llama} to determine the optimal setup for \textit{MoMug}. We conducted extensive experiments across different model scales (1B, 3B, 8B) and fine-tuned key LoRA parameters, such as LoRA rank, scaling factor (alpha), and etc, to achieve an optimal balance between efficiency and accuracy. Additionally, we explored multiple training configurations, including learning rate schedules, loss weight ($\lambda$), and etc., to mitigate overfitting and enhance model stability. Through a combination of empirical tuning and grid search, we identified the best-performing set of hyperparameters, which resulted in improved convergence speed and superior model generalization across our evaluation benchmarks.

\begin{table}[ht]
\centering
\caption{Hyper-parameters of Our Proposed LLaMA-Based Model}
\label{tab:hyperparameters_llama}
\begin{tabular}{llll}
\toprule
\multicolumn{1}{l}{\textbf{Subcategory}} & \multicolumn{1}{l}{\textbf{Hyper-parameter}} & \multicolumn{1}{l}{\textbf{Explored Value Pattern}} & \multicolumn{1}{l}{\textbf{Explanation}} \\
\hline

\multirow{6}{*}{LLaMA Model} 
& Model Sizes & \{1B, 3B, 8B\} & Different model scales used. \\
& Hidden Size & \{2048, 3072, 4096\} & Dimension of hidden layers for each model size. \\
& Num Layers & \{16, 28, 32\} & Number of transformer layers per model. \\
& Num Heads & \{32, 24, 32\} & Attention heads per model size. \\
& Sequence Length & 512 & Maximum sequence length for training. \\

\hline

\multirow{4}{*}{LoRA} 
& LoRA Rank & \{32, 64, 128, 256\} & Rank of low-rank adaptation matrix. \\
& LoRA Alpha & \{64, 128, 256, 512\} & Scaling factor for LoRA updates. \\
& LoRA Dropout & 0.1 & Dropout applied to LoRA layers. \\
& Target Layers & FF \& Attention & LoRA applied to self-attention mechanism. \\

\hline

\multirow{6}{*}{Training} 
& Learning Rate & \{1e-4, 1.5e-4\} & Initial learning rate. \\
& Weight Decay & 0.001 & Weight decay coefficient. \\
& Batch Size & 64 & Batch size per training step. \\
& Gradient Clipping & 1.0 & Max gradient norm for stabilization. \\
& Loss Weight ($\lambda$) & \{0.01, 0.005, 0.003\} & Loss weighting for different objectives. \\
& Warmup Steps & 10\% & Steps for linear warm-up. \\

\hline

\multirow{4}{*}{Optimization} 
& Optimizer & AdamW & Optimizer used for training. \\
& Beta1 & 0.9 & First moment decay rate. \\
& Beta2 & 0.95 & Second moment decay rate. \\
& Epsilon & 1e-8 & Numerical stability term. \\

\bottomrule
\end{tabular}
\end{table}

\section{Multi-Task Learning for Motion Modality Generation}
\label{app:multi-task}

To improve the generalization capability of our model, we also employed a multi-task learning strategy, systematically training the motion modality generation model by mixing motion-to-text and text-to-motion data. This approach enables the trained model to handle both tasks efficiently. However, the inference performance for each task varies depending on the mixing ratio of the two data types.

We conducted experiments using three different mixing ratios: 6:4, 7:3, and 8:2 (motion-to-text data : text-to-motion data). To analyze the impact of the mixing ratio on the model's performance, we evaluated each configuration separately on both tasks and reported relevant metrics. The results are presented in Tables~\ref{tab::data_ratio}.

\begin{table}[ht]
  \centering
  
  \caption{Ablation Study on Dataset Ratio for the Llama 1B Model on the Motion-to-Text Task.}
  \label{tab::data_ratio}
  \begin{tabular}{lcccccc}
  \toprule
   \text{Ratio(T2M: M2T)} & \textbf{RP-T3 $\uparrow$} & \textbf{MMDist $\downarrow$} & \textbf{Bleu4 $\uparrow$} & \textbf{Rouge $\uparrow$} & \textbf{Cider $\uparrow$} & \textbf{BertScore $\uparrow$} \\
  \midrule
  % \rowcolor{gray!10} 8:2(best fid)  & 0.699 & 3.572 & 12.4 & 37.4 & 26.0 & 30.7 \\
  \rowcolor{gray!10} 8:2  & 0.682 & 3.759 & 18.5 & 43.5 & 37.3 & 37.8 \\
  
  % \rowcolor{white} 7:3(best fid) & 0.743 & 3.229  & 12.8 & 38.3 & \textbf{29.1} & 32.1 \\
  \rowcolor{white} 7:3 & 0.758 & 3.103  & \textbf{20.9} & \textbf{45.7} & \textbf{45.8} & 40.3 \\
  
  % \rowcolor{gray!10} 6:4(best fid)  & 0.775 & 3.060 & 14.3 & 40.1 & 32.9 & 33.7 \\
  \rowcolor{gray!10} 6:4  & \textbf{0.799} & \textbf{2.899} & 20.3 & \textbf{45.7} & 44.8 & \textbf{40.4} \\
  \bottomrule
  \end{tabular}
\end{table}

\section{Experimental Result in KIT-ML Dataset}
\label{app:kit-result}

For the KIT-ML dataset, as shown in Table~\ref{tab:kit_ml_results}, we observed similar performance patterns. \textit{MoMug} demonstrates competitive performance across all metrics, particularly in R-Precision and MultiModal Distance, which highlights the generalizability of \textit{MoMug} across different datasets.

\begin{table*}[h]
  \caption{Quantitative results of text-based motion generation on the KIT-ML dataset}
  \label{tab:kit_ml_results}
  \scriptsize 
  \centering
  \rowcolors{2}{gray!15}{white}

  \begin{tabular}{llccccccc}
    \toprule
    \textbf{Model} & \textbf{Type} & \multicolumn{3}{c}{\textbf{R-Precision} $\mathbf{\uparrow}$} & \textbf{FID} $\mathbf{\downarrow}$ & \textbf{MultiModal Dist.} $\mathbf{\downarrow}$ & \textbf{Diversity} $\mathbf{\rightarrow}$ & \textbf{MultiModality} $\mathbf{\uparrow}$ \\
    &  & \textbf{Top1} & \textbf{Top2} & \textbf{Top3} & & & & \\
    \midrule
    Real & — 
      & $0.424^{\pm.005}$ & $0.649^{\pm.006}$ & $0.779^{\pm.006}$ & $0.031^{\pm.004}$ & $2.788^{\pm.012}$ & $11.080^{\pm.097}$ & — \\
    \midrule
    TM2T\cite{guo2022tm2tstochastictokenizedmodeling} & AR
      & $0.280^{\pm.005}$ & $0.463^{\pm.006}$ & $0.587^{\pm.005}$ & $3.599^{\pm.153}$ & $4.591^{\pm.026}$ & $9.473^{\pm.117}$ & $3.292^{\pm.081}$ \\
    %T2M\cite{} & AR
      %& $0.361^{\pm.006}$ & $0.559^{\pm.007}$ & %$0.681^{\pm.007}$ & $3.022^{\pm.107}$ & %$3.488^{\pm.028}$ & $10.720^{\pm.145}$ & %$2.052^{\pm.107}$ \\
    MDM\cite{tevet2022human} & DM
      & $0.164^{\pm.004}$ & $0.291^{\pm.004}$ & $0.396^{\pm.004}$ & $0.497^{\pm.021}$ & $9.191^{\pm.022}$ & $10.850^{\pm.109}$ & $1.907^{\pm.214}$ \\
    MotionDiffuse\cite{zhang2022motiondiffusetextdrivenhumanmotion} & DM
      & $0.417^{\pm.004}$ & $0.621^{\pm.004}$ & $0.739^{\pm.004}$ & $1.954^{\pm.062}$ & $2.958^{\pm.005}$ & $11.100^{\pm.143}$ & $0.730^{\pm.013}$ \\
    MLD\cite{zhang2023mld} & DM
      & $0.390^{\pm.008}$ & $0.609^{\pm.008}$ & $0.734^{\pm.007}$ & $0.404^{\pm.027}$ & $3.204^{\pm.027}$ & $10.800^{\pm.117}$ & $2.192^{\pm.071}$ \\
    T2M-GPT\cite{zhang2023t2mgptgeneratinghumanmotion} & AR
      & $0.416^{\pm.006}$ & $0.627^{\pm.006}$ & $0.745^{\pm.006}$ & $0.514^{\pm.029}$ & $3.007^{\pm.023}$ & $10.920^{\pm.108}$ & $1.570^{\pm.039}$ \\
    MoMask\cite{guo2023momaskgenerativemaskedmodeling} & AR
      & $\mathbf{0.433}^{\pm.007}$ & $\mathbf{0.656}^{\pm.005}$ & $\mathbf{0.781}^{\pm.005}$ & $\mathbf{0.204}^{\pm.011}$ & $\mathbf{2.779}^{\pm.022}$ & $10.711^{\pm.087}$ & $1.131^{\pm.043}$ \\
    ReMoDiffuse\cite{liu2023remodiffuse} & DM
      & $0.427^{\pm.014}$ & $0.641^{\pm.004}$ & $0.765^{\pm.055}$ & $0.155^{\pm.006}$ & $2.814^{\pm.012}$ & $10.800^{\pm.105}$ & $1.239^{\pm.028}$ \\
    AttT2M\cite{zhong2023attt2mtextdrivenhumanmotion} & AR
      & $0.413^{\pm.006}$ & $0.632^{\pm.006}$ & $0.751^{\pm.006}$ & $0.870^{\pm.039}$ & $3.039^{\pm.021}$ & $10.960^{\pm.123}$ & $2.281^{\pm.047}$ \\
    GraphMotion\cite{yin2021graph} & GNN
      & $0.429^{\pm.007}$ & $0.648^{\pm.006}$ & $0.769^{\pm.006}$ & $0.313^{\pm.013}$ & $3.076^{\pm.022}$ & $\mathbf{11.120}^{\pm.135}$ & $3.627^{\pm.113}$ \\
    \midrule
    MotionGPT\cite{jiang2024motiongpt} & AR
      & $0.340^{\pm.002}$ & $0.570^{\pm.003}$ & $0.660^{\pm.004}$ & $0.868^{\pm.032}$ & $3.721^{\pm.018}$ & $9.972^{\pm.026}$ & $2.296^{\pm.022}$ \\
    MotionGPT\cite{motiongptother2023} & AR
      & $0.366^{\pm.005}$ & $0.558^{\pm.004}$ & $0.680^{\pm.005}$ & $0.510^{\pm.016}$ & $3.527^{\pm.021}$ & $10.350^{\pm.084}$ & $2.328^{\pm.117}$ \\
    MotionLLM\cite{chen2024motionllm} & AR
      & $0.409^{\pm.006}$ & $0.624^{\pm.007}$ & $0.750^{\pm.005}$ & $0.781^{\pm.026}$ & $\mathbf{2.982}^{\pm.022}$ & $11.407^{\pm.103}$ & — \\
    MotionGPT-2\cite{wang2024motiongpt2generalpurposemotionlanguagemodel} & AR
      & $\mathbf{0.427}^{\pm.003}$ & $\mathbf{0.627}^{\pm.002}$ & $\mathbf{0.764}^{\pm.003}$ & $0.614^{\pm.005}$ & $3.164^{\pm.013}$ & $11.256^{\pm.026}$ & $\mathbf{2.357}^{\pm.022}$ \\
    \midrule
    \textit{MoMug}(t2m) & AR+DM 
      & $0.439^{\pm.004}$ & $0.352^{\pm.001}$ & $0.472^{\pm.003}$ & $\mathbf{0.390^{\pm.006}}$ & $5.170^{\pm.008}$ & $\mathbf{11.018^{\pm.102}}$ & $0.727^{\pm.021}$ \\
    \bottomrule
  \end{tabular}
\end{table*}

 \textit{MoMug} achieves a favorable FID score (0.390) and Diversity (11.018), demonstrating that our generated motions maintain a realistic quality and variation comparable to the real data. However, our R-Precision and MultiModality scores are relatively lower than some baselines. The primary reason for this limitation is the data sparsity in the KIT-ML dataset. Both diffusion models and language models typically require large datasets for effective training, but the small number of motion sequences (3911 motion sequences) in KIT-ML poses challenges for learning diverse and meaningful motion patterns. As a result, while our approach generates visually diverse motions with a good distribution quality, its retrieval-based precision and multimodal consistency suffer due to the data constraints. 

 Compared with MDM \cite{tevet2022human}, which serves as the diffusion model backbone for our method, our approach achieves better results across all metrics. Specifically, our model outperforms MDM in R-Precision (Top1, Top2, Top3), FID, Diversity, and MultiModality, demonstrating significant improvements in both motion quality and generation diversity. This suggests that our enhancements effectively address the limitations of the diffusion-based baseline, leading to higher-quality motion generation with improved fidelity and diversity. These results highlight the effectiveness of  \textit{MoMug} in refining diffusion-based motion synthesis, even under the data constraints of KIT-ML.

\section{Training and Inference Time}
\label{app:train-inference-time}

\begin{table}[h]
    \centering
    \caption{Training and inference times on HumanML3D~\cite{guo2022generating} for different model sizes along with the required GPU.}
    \label{tab:training_inference_times}
    \renewcommand{\arraystretch}{1.2} % Adjust row height
    \setlength{\tabcolsep}{8pt} % Adjust column spacing
    \begin{tabular}{lccc}
        \toprule
        \textbf{Model Size} & \textbf{Training Time} & \textbf{Inference Time (Per Batch on H100×1)} & \textbf{Training GPU} \\
        \midrule
        1B  & Around 65 hours  & 4.68 s (batch size 32)  & Nvidia H100×1 \\
        3B  & Around 125 hours  & 10.6 s (batch size 32)  & Nvidia H200×1 \\
        8B  & Around 55 hours  & 22.6 s (batch size 32) & Nvidia H100×8 \\
        \bottomrule
    \end{tabular}
\end{table}

Table~\ref{tab:training_inference_times} summarizes the training and inference times for \textit{MoMug} of different sizes (1B, 3B, and 8B) and the corresponding GPU configurations used for training. As the model size increases, both the training and inference times are affected, reflecting the growing computational and memory requirements. For training, different GPU configurations were utilized—Nvidia H100×1 for the 1B model, H200×1 for the 3B model, and H100×8 for the 8B model—to efficiently handle the varying resource demands. However, all inference time measurements were conducted on a single H100 GPU (H100×1) regardless of model size to provide a standardized comparison.
These results underscore the trade-off between model complexity and computational cost, as well as the importance of selecting an appropriate GPU for each model size to achieve optimal performance.

\section{Visualized Result}
\label{app:Visualized-result}

The qualitative results presented in Table~\ref{tab:motion_comparison} demonstrate that our proposed MoMug model generates more natural and coherent motion sequences compared to the baselines MDM \cite{tevet2022human} and MotionGPT \cite{jiang2024motiongpt}. Specifically, MoMug produces smoother transitions, better pose consistency, and fewer unnatural artifacts. In contrast, MDM often exhibits unstable motion trajectories, while MotionGPT tends to generate overlapping or unrealistic poses due to its reliance on autoregressive token prediction. The ability of MoMug to generate high-quality motion sequences highlights the effectiveness of combining LLM-based text encoding with motion diffusion modeling, making it a superior approach for text-to-motion generation.

The results in Table~\ref{tab:text_generation_comparison} clearly demonstrate the strength of our motion-to-text capability. Our proposed \textit{MoMug} model consistently produces detailed and accurate textual descriptions that capture the subtle nuances of the input motion data. Unlike the baseline methods, which often generate imprecise or overlapping narratives, \textit{MoMug} provides clear, coherent, and natural language outputs. This superior performance highlights the effectiveness of our unified motion-language representation, underscoring its ability to bridge the gap between visual motion data and linguistic expression.

\begin{table*}[ht]
    \centering
    \renewcommand{\arraystretch}{1.5} % Adjust row height
    \setlength{\tabcolsep}{2pt} % Reduce column spacing slightly
    
    \resizebox{0.95\textwidth}{!}{ % Resize table to fit within text width
    \begin{tabular}{>{\centering\arraybackslash}m{2.5cm} >{\raggedright\arraybackslash}m{4.5cm} >{\raggedright\arraybackslash}m{4.5cm} >{\raggedright\arraybackslash}m{4.5cm}} % Slightly reduce column width
        \toprule
        \textbf{Method} & \textbf{Motion Description 1} \rule{0pt}{2ex} 
        & \textbf{Motion Description 2} \rule{0pt}{2ex}
        & \textbf{Motion Description 3} \rule{0pt}{2ex} \\ 
        \midrule
        & \makecell[tl]{A person is running on \\ the spot, then turns left \\ and jabs with both arms,\\ then turns right and con- \\tinues running on the spot.} 
        & \makecell[tl]{A person walks to the right, \\ makes a u-turn clockwise, \\ and returns to the left of \\ their initial position facing \\ away.}
        & \makecell[tl]{A dance that symbolizes \\ the feeling of the figure \\ walking from the bottom \\right to the top left of the \\ square, and bending down \\as if picking something up \\twice.} \\
        \midrule
        \multirow{1}{=}{\centering MDM \cite{tevet2022human}} 
        & \includegraphics[width=\linewidth]{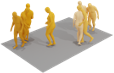} 
        & \includegraphics[width=\linewidth]{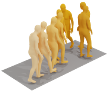} 
        & \includegraphics[width=\linewidth]{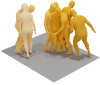} \\
        \midrule
        \multirow{1}{=}{\centering MotionGPT \cite{jiang2024motiongpt}}  
        & \includegraphics[width=\linewidth]{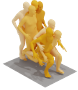} 
        & \includegraphics[width=\linewidth]{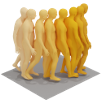} 
        & \includegraphics[width=\linewidth]{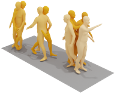} \\
        \midrule
        \multirow{1}{=}{\centering Our \textit{MoMug}}  
        & \includegraphics[width=\linewidth]{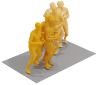} 
        & \includegraphics[width=\linewidth]{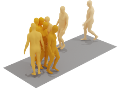} 
        & \includegraphics[width=0.6\linewidth]{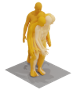} \\
        \bottomrule
    \end{tabular}
    } % End resizebox

    \caption{Comparison of motion generation results across different methods. Each row represents a model (MDM, MotionGPT, MoMug), and each column corresponds to a different text-based motion description.}
    \label{tab:motion_comparison}
\end{table*}

\begin{table*}[ht]
    \centering
    % Adjust row height
    \renewcommand{\arraystretch}{1} 
    % Reduce column spacing slightly
    \setlength{\tabcolsep}{2pt} 
    
    % Resize table to fit within text width
    \resizebox{0.95\textwidth}{!}{ 
    \begin{tabular}{%
        >{\centering\arraybackslash}m{2.5cm}  % Method column
        >{\centering\arraybackslash}m{3.5cm}  % Motion 1 column
        >{\centering\arraybackslash}m{3.5cm}  % Motion 2 column
        >{\centering\arraybackslash}m{3.5cm}  % Motion 3 column
        >{\centering\arraybackslash}m{3.5cm}  % Motion 4 column
    }
        \toprule
        \textbf{Method} & \textbf{Motion Data 1} & \textbf{Motion Data 2} & \textbf{Motion Data 3} & \textbf{Motion Data 4} \\
        \midrule
        
        % -- First row: Input Motion (images) --
        \textbf{Input Motion} 
        & \includegraphics[width=\linewidth]{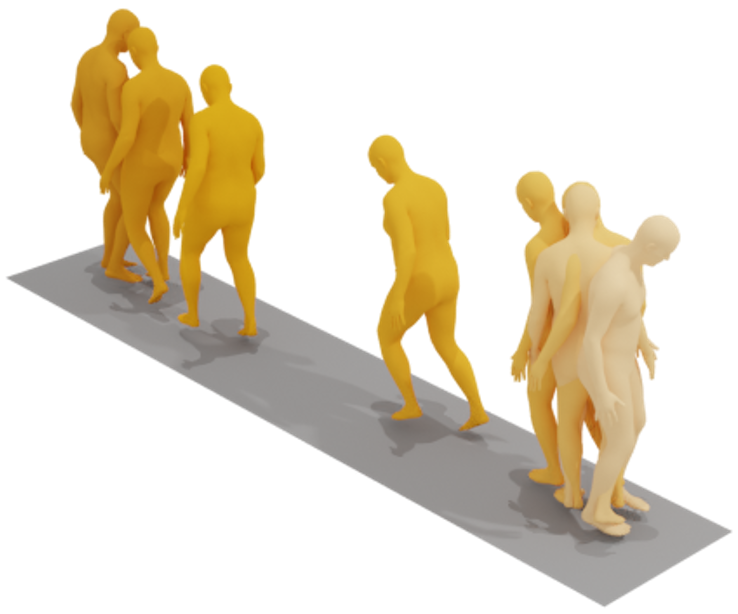}
        & \includegraphics[width=0.7\linewidth]{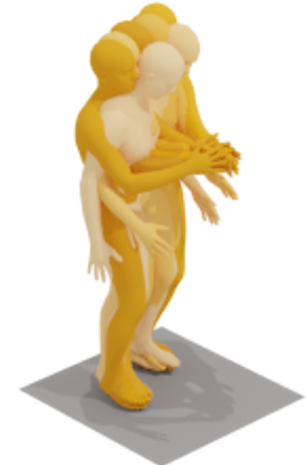}
        & \includegraphics[width=0.5\linewidth]{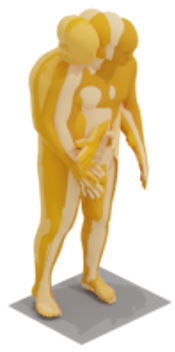}
        & \includegraphics[width=\linewidth]{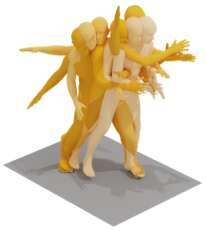}
        \\
        \midrule
        
        % -- Next row(s): Method A's text descriptions --
        Real
        & {\scriptsize 
            \makecell[tl]{%
                1) the person is walking forward \\[0.1ex]
                and turns around like a monster. \\[0.1ex]
                2) a person moves forward  \\[0.1ex]
                sluggishly then runs in another \\[0.1ex]
                direction.\\[0.1ex]
                3) a person takes a few steps \\[0.1ex]
                forwards, then turns around and \\[0.1ex]
                tries to jog.
            }
        }
        & {\scriptsize 
            \makecell[tl]{%
                1) excited person starts with \\[0.1ex]
                arms out and bounces from \\[0.1ex]
                out and bounces from one foot\\[0.1ex]
                to the other while clapping.\\[0.1ex]
                2) a person moves back and \\[0.1ex]
                forth to loosen up their \\[0.1ex]
                muscles, and then claps \\[0.1ex]
                their hands to cheer \\[0.1ex]
                someone on.\\[0.1ex]
                3) a figure waves its arms, \\[0.1ex]
                like it's stretching before \\[0.1ex]
                working out, then claps its\\[0.1ex]
                hands.
            }
        }
        & {\scriptsize 
            \makecell[tl]{%
                1) a person lifts something \\[0.1ex]
                up and to the side to look \\[0.1ex]
                under it and puts the item\\[0.1ex]
                back down.\\[0.1ex]
                2) a person picks up two \\[0.1ex]
                objects, pours something \\[0.1ex]
                from the left object into \\[0.1ex]
                the right object, and then\\[0.1ex]
                sets them back down.\\[0.1ex]
                3) a person lifts object on \\[0.1ex]
                to its side from left to \\[0.1ex]
                right.\\[0.1ex]
            }
        }       
        & {\scriptsize 
            \makecell[tl]{%
                1) a person is pitching a baseball.\\[0.1ex]
                2) a person moves their right hand \\[0.1ex]
                that starts in front of their body\\[0.1ex]
                with their other hand, in a counter\\[0.1ex]
                clockwise circle to the right of\\[0.1ex]
                their body and when the arm \\[0.1ex]
                reaches back to the front of the \\[0.1ex]
                circle the left hand in brought \\[0.1ex]
                to meet the left hand. this happens \\[0.1ex]
                again as if the person is throwing \\[0.1ex]
                a ball with their right hand.\\[0.1ex]
                3) the person takes a step back\\[0.1ex]
                with it's right foot, raising its\\[0.1ex]
                right hand in a throwing motion\\[0.1ex]
                similar to a baseball pitcher. \\[0.1ex]
            }
        }
        \\
        \midrule
        
        % -- Next row(s): Method B's text descriptions --
        MotionGPT\cite{jiang2024motiongpt}
        & {\scriptsize 
            \makecell[tl]{%
            a man staggers forward, turns \\[0.1ex]
            counterclockwise, then jogs \\[0.1ex]
            back to starting point.
            }
        }
         & {\scriptsize 
            \makecell[tl]{%
            the person is shaking out \\[0.1ex]
            both arms. \\[0.1ex]
            }
        }
        & {\scriptsize 
            \makecell[tl]{%
            a person lifts an object up and \\[0.1ex]
            tilts it to the right side before\\[0.1ex]
            returning it to it's original location.\\[0.1ex]
            }
        }
         & {\scriptsize 
            \makecell[tl]{%
            a person bring both hands together \\[0.1ex]
            then throws with their right and \\[0.1ex]
            then throws again.\\[0.1ex]
            }
        }
        \\
        \midrule
        
        % -- Next row(s): Method C's text descriptions --
       Our \textit{MoMug} 
        & {\scriptsize 
            \makecell[tl]{%
                a person walks forward, turns \\[0.1ex]
                around, and then jogs back.\\[0.1ex]
            }
        } 
        & {\scriptsize 
            \makecell[tl]{%
                a person is moving their hands \\[0.1ex]
                around in a circle. \\[0.1ex]
        }
        }
        & {\scriptsize 
            \makecell[tl]{%
                a person picks up an object, \\[0.1ex]
                tilts it to the left, then \\[0.1ex]
                places it down.\\[0.1ex]
        }
        }
        & {\scriptsize 
            \makecell[tl]{%
               a person throws a football  \\[0.1ex]
               with their right hand. \\[0.1ex]
        }
        }
        \\
        \bottomrule
    \end{tabular}
    } % End resizebox

    \caption{Comparison of text generation for different input motions (four in total) across various methods.}
    \label{tab:text_generation_comparison}
\end{table*}

\end{document}